\documentclass[
]{ceurart}

\usepackage{commath}
\usepackage{graphics}
\usepackage{siunitx}

\begin{document}

\copyrightyear{2021}
\copyrightclause{Copyright for this paper by its authors.
  Use permitted under Creative Commons License Attribution 4.0
  International (CC BY 4.0).}

\conference{De-Factify: Workshop on Multimodal Fact Checking and Hate Speech Detection, co-located with AAAI 2022. 2022
Vancouver, Canada}

\title{Logically at Factify 2022: Multimodal Fact Verification}

\author{Jie Gao}[%
orcid=0000-0002-3610-8748,
email=jie@logically.ai
]

\author{Hella-Franziska Hoffmann}[%
email=hella.h@logically.ai
]

\author{Stylianos Oikonomou}[%
email=stylianos@logically.ai
]

\author{David Kiskovski}[%
email=david.k@logically.ai
]

\author{Anil Bandhakavi}[%
email=anil@logically.ai,
url=https://www.logically.ai/team/leadership/anil-bandhakavi,
]

\address{Brookfoot Mills, Brookfoot Industrial Estate, Brighouse, HD6 2RW, United Kingdom}


%

\begin{abstract}
This paper describes our participant system for the multi-modal fact verification (Factify) challenge at AAAI 2022. Despite the recent advance in text-based verification techniques and large pre-trained multimodal models cross vision and language, very limited work has been done in applying multimodal techniques to automate fact checking processes, particularly considering the increasing prevalence of claims and fake news about images and videos on social media. In our work, the challenge is treated as a multimodal entailment task and framed as multi-class classification. Two baseline approaches are proposed and explored including an ensemble model (combining two uni-modal models) and a multi-modal attention network (modeling the interaction between image and text pair from claim and evidence document). We conduct several experiments investigating and benchmarking different SoTA pre-trained transformers and vision models in this work. Our best model is ranked first on the leaderboard and obtains a weighted average F-measure of 0.77 on both validation and test set. Exploratory analysis is also carried out on the Factify data set and uncovers salient patterns and issues (e.g. word overlap, visual entailment correlation, source bias) that motivates our hypothesis. Finally, we highlight challenges of the task and multimodal dataset for future research.


\end{abstract}


\maketitle

\section{Introduction}
\noindent Rapidly growing volume of misinformation and fake news have become a pressing challenge and cause severe consequences on society. Significant joint efforts have been undertaken by a wide range of parties (represented by journalists, researchers, independent fact checkers) to protect communities from false information. It has never been more important to have a versatile ecosystem to scale up and speed up fact checking against misinformation using technology, which can be broadly categorised into claim detection and claim validation \cite{zeng2021automated}. The former technique is to support fact checkers in content prioritisation through assessing check-worthiness, and the latter one is to automate the process of evidence retrieval from large knowledge bases and performing veracity prediction of the detected claims in order to assist manual fact checking tasks. Claim matching \cite{kazemi2021claim} is another emerging trend, addressing the need for timely identification of previously fact-checked claims. Prior efforts focused mostly on text from news media articles and English language. In recent years, with the advance in user-generated content and increasingly polarized social platforms, the challenges of fact checking have increasingly become multilingual and multimodal which have been pervasive in user-generated multimedia content \cite{jang2019fake}. As a consequence, many new problems arise, typically false context, false connections or misleading content \cite{nakamura2019r,zlatkova2019fact,elhadad2020detecting,alam2021survey,sun2021inconsistency}. Another understudied process (known as amplification) is leveraged by coordinated disinformation campaigns \cite{kazemi2021claim}. It deliberately spreads large volumes of repeated claims in many different ways, in order to stimulate unintentional spread as false rumors \cite{sun2021inconsistency}. Thus, there is an imperative need to develop algorithms to group same claims resided in various multimodal context and automate the verification process at scale.


Compared to text-based fact checking, multimodal verification is an under-explored area of research. Image and text both contain rich information but reside in heterogeneous modalities. Comparing representation learning within the same modality, cross-modal architectures need not only learn the features for image and text to express their respective content but importantly capture a measure for cross-modal \textbf{semantic integrity}~\cite{sun2021inconsistency,muller2021multimodal}. We study \textbf{multimodal entailment} in this paper. As a newly introduced subtask, it poses extra critical challenges.



Simple image similarity cannot resolve fine-grained images differences and perform poorly for adversarial images \cite{moosavi2016deepfool}, SAR images\cite{deledalle2014exploiting}, etc. To exemplify this challenge, two pairs of claim and document from insufficient multimodal samples in Factify dataset \cite{mishraFactify2021} are presented in Table~\ref{tbl:insufficient_multimodal_images}. The first sample shows two separate images of a politician taken in direct point-of-view, sitting at the exact same table, in the exact same room, giving a televised speech on different days for different issues. In both images, the politician is wearing a suit, in one image black, and in the other white. In this case, the images are likely to yield high similarity with respect to their content, but they should be considered different images and representative of different contextual information. The second sample presents two images of the same nature, where the politician is wearing the same white suit and ear plugs but with a news broadcasting logo overlaid on the upper right corner of claim image, and the other document image with no news channel logo visible. The main discrepancy is presented between text and image in document which reports that the politician is wearing a white mask during the video conference. Therefore, although the document text provide supporting evidence about the claim but the image is missing important context information. On the contrary, the sample in Table~\ref{tbl:support_multimodal_images} presents two images having low content overlap but the document image corresponds to its textual content that supports the politician death information as presented in claim image. Thus, the right image should be considered as supporting image and representative of same information contextually with corresponded claim image.

\begin{table}[h!]
  \centering
  \linespread{0.5}
  \rmfamily
  \begin{tabular}{ | m{4.4cm} | m{4.4cm} | }  \hline
    \multicolumn{1}{|c|}{\centering{\small Claim}} & 
     \multicolumn{1}{c|}{\centering{Document}}\\ \hline
    \begin{minipage}{.30\textwidth}
        \begin{center}
      \includegraphics[width=\linewidth, height=30mm]{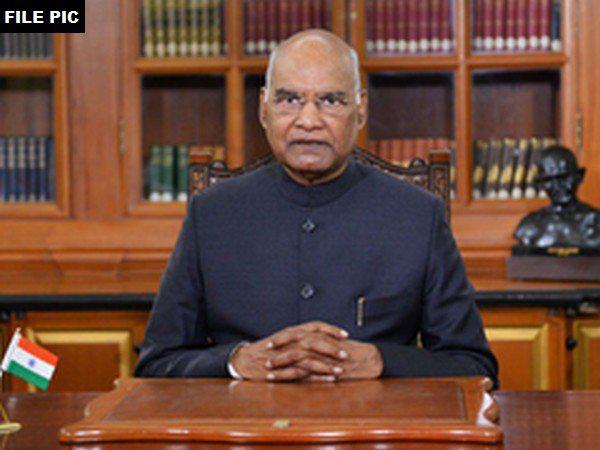}
      \end{center}
    \end{minipage}
    {\tiny In the demise of Union Minister Ram Vilas Paswan, the nation has lost a visionary leader. He was among the most active and longest-serving members of parliament...}
    &
    \begin{minipage}{.30\textwidth}
      \begin{center}
      \includegraphics[width=\linewidth, height=30mm]{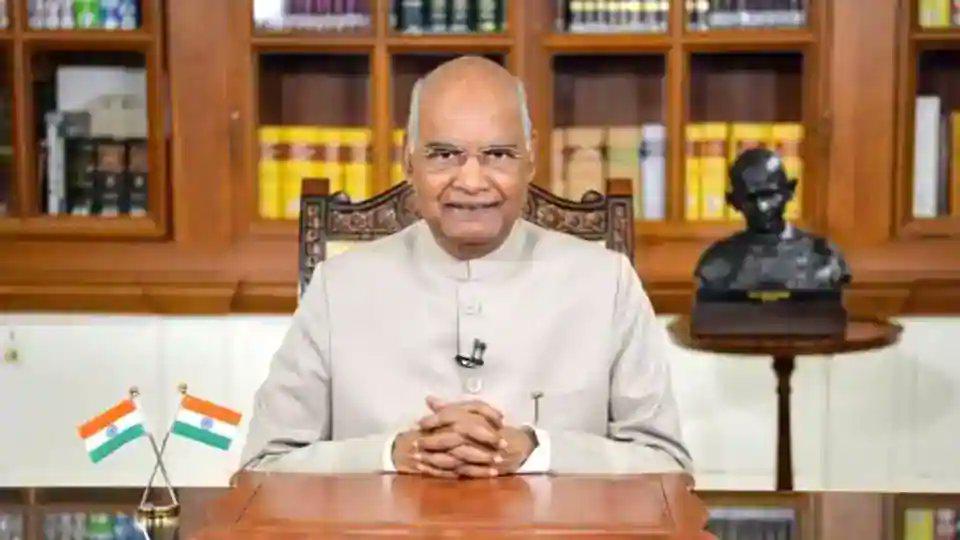}
      \end{center}
    \end{minipage}
    {\tiny ... President Ram Nath Kovind said on Wednesday, ... Addressing the fourth annual convocation of the Jawaharlal Nehru University, he said said Indian scholars of today ...}
    \\\hline
    \begin{minipage}{.30\textwidth}
      \includegraphics[width=\linewidth, height=30mm]{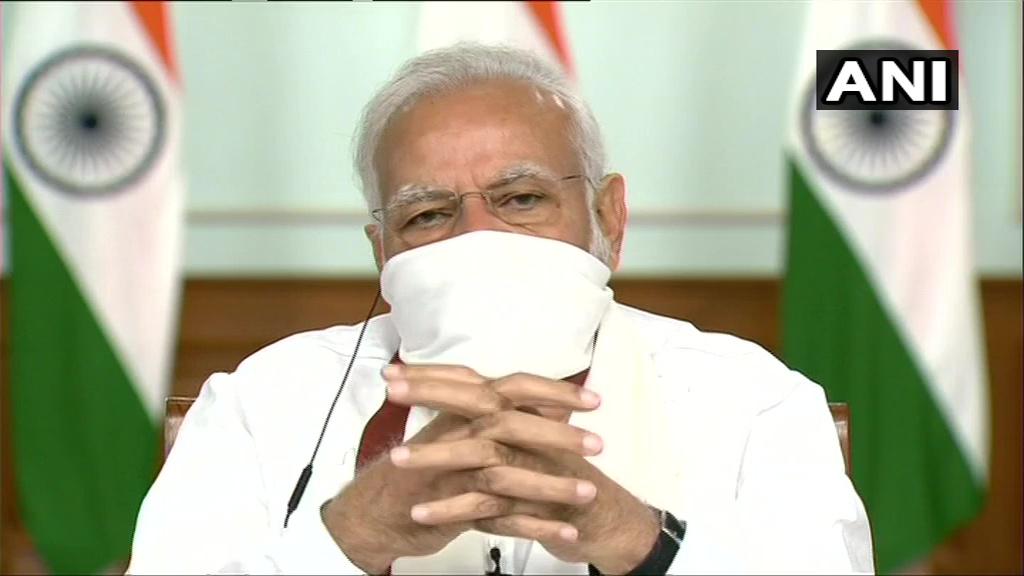}
    \end{minipage}
    {\tiny Prime Minister Narendra Modi holds a meeting via video-conferencing with the Chief Ministers over \#COVID19...}
    &
    \begin{minipage}{.30\textwidth}
      \includegraphics[width=\linewidth, height=30mm]{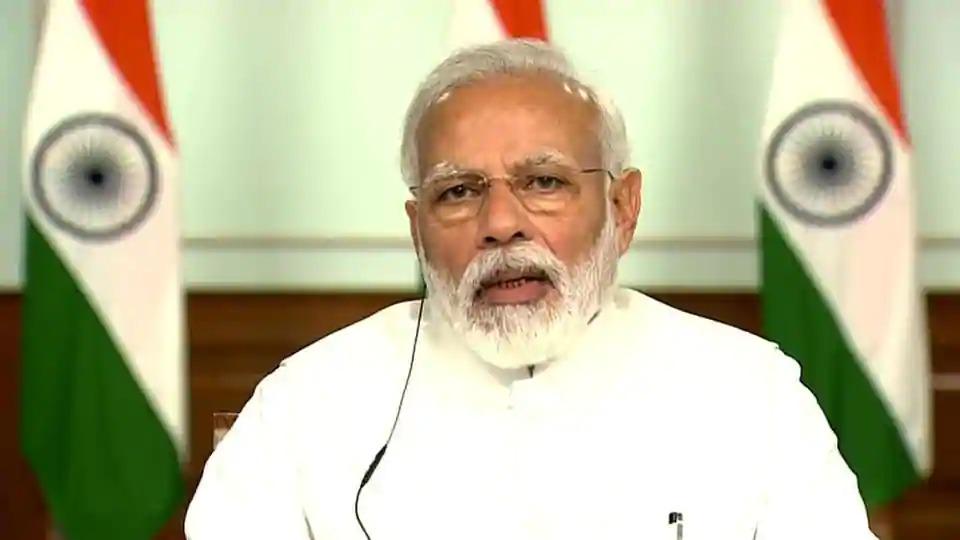}
    \end{minipage}
    {\tiny ... Prime Minister Narendra Modi on Saturday held a video conference with  ... showed Modi wearing a white mask during the interaction ...}
    \\ \hline
  \end{tabular}
  \caption{\footnotesize Insufficient Multimodal examples in Factify dataset. Claim image+text(left), Document image+text(right)}\label{tbl:insufficient_multimodal_images}
\end{table}

\begin{table}[h!]
  \centering
  \linespread{0.5}
  \rmfamily
  \begin{tabular}{ | m{4.4cm} | m{4.4cm} | }  \hline
    \multicolumn{1}{|c|}{\centering{Claim}} & 
     \multicolumn{1}{c|}{\centering{Document}}\\ \hline
    \begin{minipage}{.30\textwidth}
      \includegraphics[width=\linewidth, height=30mm]{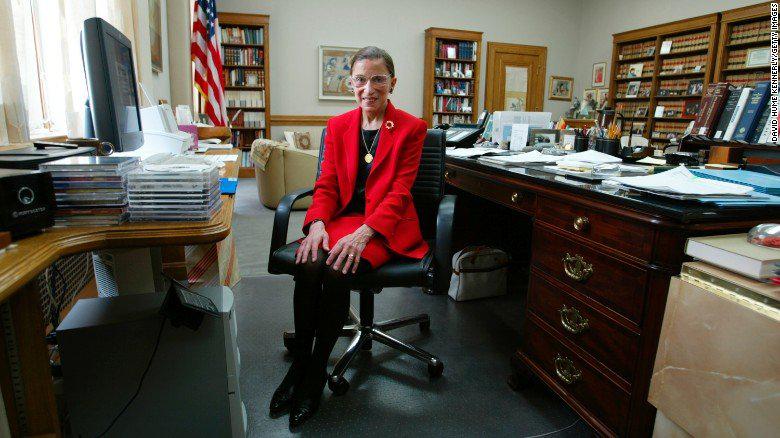}
    \end{minipage}
    {\tiny Here's a look back at the life and legacy of Ruth Bader Ginsburg, the second woman to serve on the US Supreme Court, in photos. Ginsburg died Friday due to  ...}
    &
    \begin{minipage}{.30\textwidth}
      \includegraphics[width=\linewidth, height=30mm]{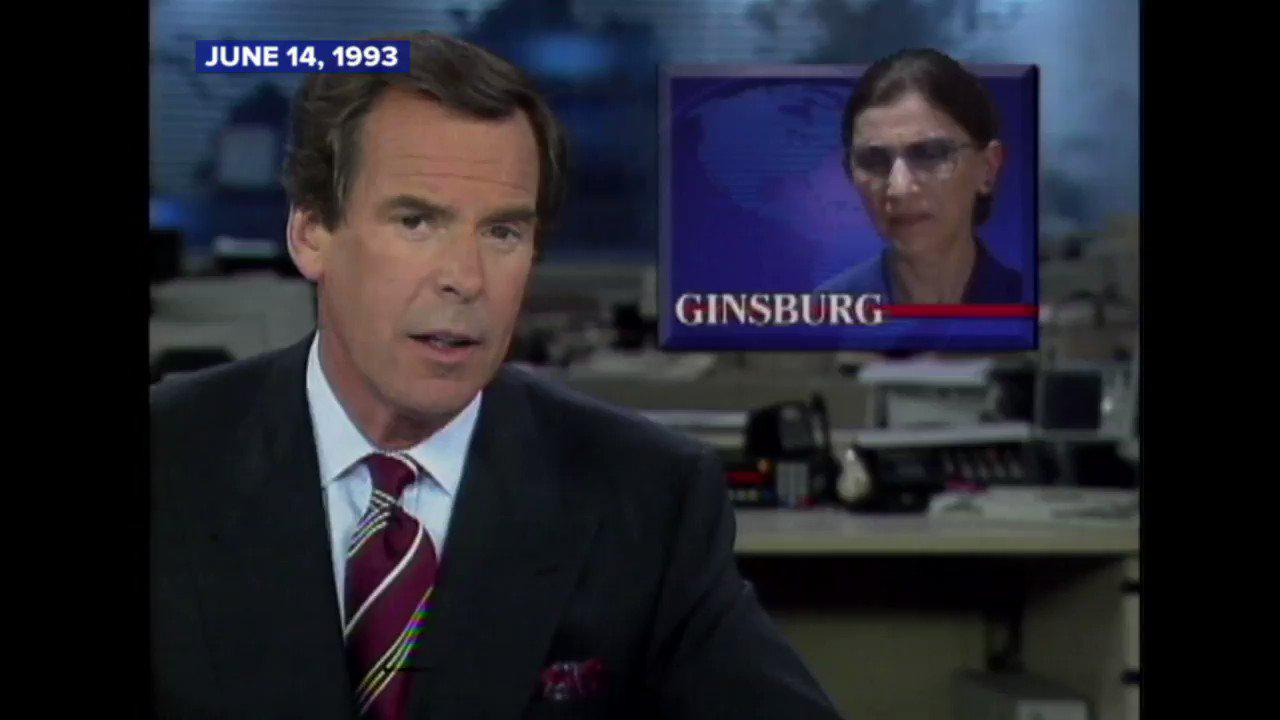}
    \end{minipage}
    {\tiny She was appointed to the Supreme Court by Bill Clinton in 1993.  Remembering Supreme Court Justice Ruth Bader ... has died at the age of 87. ... lost a cherished colleague,"" Chief Justice John Roberts said ...}
    \\\hline
  \end{tabular}
  \caption{\footnotesize Support Multimodal examples in Factify dataset. Claim image+text(left), Document image+text(right)}\label{tbl:support_multimodal_images}
\end{table}

Relying on visual similarity analysis alone for multimodal fact verification is naturally prone to false positives, because images related to branding and advertisements (e.g., the “breaking news” image or a company’s logo) are often reused. This may cause erroneous detection when there is no real connection between them other than the reuse of a generic image. The problem becomes more complex with images exploited in disinformation on social media.


Our work in this competition is in response to current online misinformation multimodal issue and has focused on solving the above challenges. Two different algorithms are designed for the task that is framed as a multimodal entailment prediction problem following two different frameworks, including an ensemble learning and an end-to-end attention network. The ensemble model approach is implemented with a decision tree classifier that combines predictions of two uni-modal models with a few data-specific heuristic features. Two uni-modal models are implemented including a 3-way text entailment model based on a State-of-the-Art (SoTA) pre-trained transformer language architectures fine-tuned on the task dataset, and a pre-trained CNN model (ResNet-50) for image similarity. A SoTA multimodal attention network for 5-way end-to-end entailment classification is implemented as an alternative solution in attempt to infer combinatorial entailment relation by combining representation of language and vision. Globe-level multimodal interactions are modeled with a popular multi-branch attention network framework in order to fuse multimodal information. Strong baselines are implemented for both 3-way and 5-way text entailment models to prove the advantage of our proposed methods. Exploratory data analysis and bias test experiments are conducted to understand the potential data issues and present the challenge of creating high-quality multimodal datasets for the real-world problem. Best results from the ensemble model were submitted for the competition.


In the remainder of this paper, we firstly present a brief overview of related work (Section~\ref{sec:rel}), then task definition and our proposed methods in details (Section~\ref{sec:method}) followed by experiments on the task dataset (Section~\ref{sec:experiment}). Exploratory data analysis is elaborated in Section~\ref{sec:data}.  Finally, the results discussion (including 3-way models and 5-way models) and conclusion are provided in Sections~\ref{sec:results} and \ref{sec:conclude} respectively.



\section{Related Work}\label{sec:rel}

\textbf{Text Entailment}
Recognising Textual Entailment (RTE) is earliest and most related work to our Factify challenge that aims to determine an inferential relationship between natural language hypothesis and premise. On the basis of a given sentence pair, the task is to predict 3-way labels including \textit{Support}, \textit{Refute} or \textit{NotEnoughInfo}. Well-known shared tasks include FEVER~\cite{thorne2018fever} and SCIVER~\cite{wadden2020fact}, which advanced RTE research for claim validation in recent years. This line of work performs different forms of evidence retrieval and then applies claim validation based on that evidence. In contract, evidence retrieval is not required in the Factify task (although the practice of sentence retrieval~\cite{zhou2019gear,chen2016enhanced} as a classic NLI problem for long document text in Factify data can be considered as good practice and applicable). Stance detection is another direction of work supported by shared tasks such as UKP Snopes~\cite{hanselowski2019richly}, Semeval-2017 Rumoureval \cite{derczynski2017semeval}), and has also been exploited for RTE by retrieving texts relevant to a claim or story, determining the stance of those text afterwards so as to ultimately predict the veracity of a given claim. The common practice of RTE for claim verification~\cite{zeng2021automated} is also incorporated in our ensemble model (as one of the proposed solutions) and treated as a three-way text classification task on text data. Sentence retrieval for evidence aggregation and stance detection are not exploited in this work.

\textbf{Multi/cross-modal representation learning} In the field of multimodal reasoning and matching, the success of attention mechanism in the NLP community motivated computer vision techniques to shift from traditional twin network (typically with Siamese nets\cite{gu2018look,wang2018joint,nam2017dual}) to pre-train models in multimodal settings for wide range of downstream tasks, such as visual question answering (VQA), visual reasoning and image captioning. Similar to BERT\cite{devlin2018bert}, the recent approach is to use a single transformer architecture to jointly encode text and image such as VisualBERT~\cite{li2019visualbert}, Uniter~\cite{chen2019uniter} and VL-BERT~\cite{su2019vl}. Alternatively, ViLBERT~\cite{lu2019vilbert} and LXMERT \cite{tan2019lxmert} introduced the two-stream architecture, where two transformers are applied to images and text independently, which is fused by a third transformer in a later stage. These models typically rely on region-based image features extracted by pre-trained object detectors based on commonly used two-staged detectors (typically Faster R-CNN model\cite{ren2015faster} or its extension Mask-RCNN \cite{he2017mask}), or single-stage detectors (typically SSD and YOLO V3 \cite{adarsh2020yolo}) or anchor-free detectors(e.g., \cite{yu2021pp}). Another directions are patch embedding\cite{tolstikhin2021mlp,dosovitskiy2020image,yuan2021tokens,chu2021conditional,liu2021survey}. This direction of work directly operates on patches (as a sequence of tokens with fixed length). Image patches and text token embeddings are fed into a transformer or self-attention model to learn fused cross-modal attention. The great progress of these recently developed models can be witnessed on the leader boards of various tasks without using ensembling such as VQA, GAQ\cite{hudson2019gqa}, NLVR2~\cite{suhr2018corpus}, which can mainly be attributed to the availability of large scale weakly correlated multimodal data (typically captioned images or video clips and accompanying subtitles \cite{desai2021redcaps}) that can be utilised to learn cross-modal representation by contrastive learning \cite{hadsell2006dimensionality}.  However, existing pre-trained models use mostly \textbf{scene-limited image-text pairs} with short and relatively simple descriptive captions for images, while ignoring richer uni-modal text data and domain-specific information. This leads to the difficulties in comprehending long paragraphs than short text\cite{schneider2021towards}. Thus, most such task (e.g., VQA, VAC, image retrieval) still need additional fusion layer to model interaction between visual and linguistic content. Moreover, limited ground truth information forces many tasks to use evaluation metrics based on binary relevance. Different from most of current cross-modal reasoning tasks, our work aims to model long sequence text and images for claim verification. In contrast to these multimodal architectures, we utilize the individual components from uni-modal pre-trained architectures. The equivalent architecture is employed by \cite{kiela2019supervised} for image-text pair interaction, however, we exploited richer cross-modal interaction among vision and text pairs. Inspired by the practice in \cite{lee2018stacked}, a stacked attention mechanism is exploited in our solution for cross-modal matching by inferring the latent language-vision alignments at a global level. Recent advance in fine-grained cross-modal representation learning approaches for region-word correspondence are not exploited in this work.

\textbf{Relevance matching technique} Relevance matching (RM) is the core problem of information retrieval (IR) and has also been applied for the detecting the entailment relation\cite{liu2016deep,de2006learning} by computing the best alignment of hypothesis to premise based on local and global interactions. \citet{vo2020facts} exploited a neural ranking model using textual and visual modalities to match multimodal claim with fact-checked information. Their model unifies textual and visual interaction between a claim and a collection of candidate articles, while Factify task aims to match a claim with one given candidate document. In our proposed solution, we extended the matching module introduced in \citet{vo2020facts} in order to better handle text with vary length.

\textbf{Visual Entailment} Visual Entailment (VE) \cite{xie2019visual} is a variant of traditional RTE task that consists of image-sentence pairs whereby a premise is defined by an image, rather than a natural language sentence. The problem that VE is trying to solve is to reason about the relationship between an image as premise P$_{image}$ and a text as hypothesis H$_{text}$. This is different from Factify task that aims to reason about the multimodal relationship between a hypothesis and premise pair of both textual and visual content with respected to five categories. Moreover, the premise text is of vary lengths rather than short hypothesis sentences in SNLI-VE dataset.

\section{Methodology} \label{sec:method}



\subsection{Problem Statement}\label{sec:problem_statement}

We frame the Factify task as a problem of multimodal entailment, which is to reason about the relationship of a multimodal claim as hypothesis and a multimodal document as premise.

Specifically, given a multimodal hypothesis (e.g. tweet) denoted by ${Q = q_{image} + q_{text}}$  and a document (typically one or more fact-checking articles) denoted by ${D = d_{image} + d_{text}}$, both of which contain one image and a text, we aim to derive a function ${f(q,d)}$ that infers their entailment of five categories (\textit{"Support\_Multimodal"}, \textit{"Support\_Text"}, \textit{"Insufficient\_Multimodal"}, \textit{"Insufficient\_Text"}, \textit{"Refute"}). The label assignment is based on the relationship conveyed by $(D,Q)$,

\begin{itemize}
\item \textbf{Support\_Multimodal} holds if there is enough evidence in ${d_{text}}$ to conclude that ${q_{text}}$ is true and ${d_{image}}$ is relevant to ${q_{image}}$ and ${q_{text}}$ in the same information context,
\item \textbf{Support\_Text} holds if there is enough evidence in ${d_{text}}$ to conclude that ${q_{text}}$ is true but ${d_{image}}$ is irrelevant to ${q_{image}}$ and does not provide supplemental information for ${q_{text}}$,
\item \textbf{Insufficient\_Multimodal} holds if the evidence in ${d_{text}}$ is insufficient to draw a conclusion about ${q_{text}}$ but ${d_{image}}$ is relevant to ${q_{image}}$ and ${q_{text}}$ in the same information context,
\item \textbf{Insufficient\_Text} holds if the evidence in ${d_{text}}$ is insufficient to draw a conclusion about ${q_{text}}$ and ${d_{image}}$ is irrelevant to ${q_{image}}$ and does not provide supplemental information for ${q_{text}}$, 
\item otherwise, the relationship is \textbf{Refute}, implying that there is enough evidence in ${d_{text}}$ to conclude that $q_{text}$ is false and ${d_{image}}$ is irrelevant to Q of both visual and text content.
\end{itemize}

Additional details of the task definition can be referred in \cite{mishraFactify2021}.


\subsection{3-way Text Entailment}\label{sec:threeway_text_ent}


Recognizing entailment in natural language is a straightforward application for fact verification. In this section, we aim to study how well a SoTA textual entailment model can be fine-tuned on the textual data pairs in Factify data set and then used in a three-way RTE task. This allows us to further assess and benchmark our proposed solution of combining two uni-modal models prediction into an ensemble model for final 5-way multimodal entailment prediction.

\textbf{Pretrained transformer fine-tuning}: Pretrained transformer models
\cite{devlin2018bert,liu2019roberta,radford2019language} have become the de facto models for a wide range of NLP tasks and provide SoTA results for RTE tasks\cite{gao2020making}. More specifically, in this work, we attempt to investigate \textit{how a pretrained model can learn to conduct RTE on the given dataset without utilizing hidden dataset bias},  \textit{how efficiently it can learn and generalise on the test set}. The problem is different from existing benchmark datasets (MultiNLI\cite{williams2018multi}, SNLI\cite{bowman2015large}, Adversarial-NLI\cite{nie2019adversarial}) mostly consisting of short sentences. Fact verification task requires to apply natural language inference (NLI) on long paragraphs or articles. As mentioned above, to simplify the problem, the practice of evidence sentence selection \cite{thorne2021evidence} that are commonly adopted in SoTA evidence-aware fact checking system are not included in our study. Thus, supported maximum sequence length and optimum document context size are two of key factors to be considered. 

Transformer-based models, such as BERT, have been one of the most successful deep learning models for NLP, but one of their core limitations is the quadratic dependency (mainly in terms of memory) on the sequence length due to their full attention mechanism. For that reason, Google's BigBird model is selected in this study, which is one of the most successful long-sequence transformers that supports sequence length of 4000 tokens. To deal with the limitations that other models face, BigBird uses a sparse attention mechanism that reduces the quadratic dependency to linear\cite{zaheer2021big}. That means that it can handle sequences of length up to 8x of what was previously possible using similar hardware. As a consequence of the capability to handle longer context, these models drastically improve performance on various NLP tasks such as claim verification for long sequences\cite{stammbach2021evidence}. The model is fine-tuned as pair-wise classification task on re-purposed data samples converted from 5-way categories to three-way categories. Formally, given a pair of text sequence (denoted as $q_{text}$ and $d_{text}$) from $Q$ and $D$, we aim to fine-tune a pre-trained model to map any pair of $(Q,D)$ to a label $y$, that determines the pre-defined textual entailment relationship ("support", "refute", "insufficient") between $q_{text}$ and $d_{text}$. The problem is treated as a supervised learning task and a set of training examples in the form of $(Q,A,y)$ is given. $[SEP]$ is added as a separator between the two inputs in pre-processing and a softmax classifier is added on top of the $[CLS]$ token in the last layer to make predictions.



\textbf{MatchPyramid}: In contrary to computationally expensive transformer models, we propose a simple baseline text entailment model with relevance matching technique. Intuitively, an article may be relevant to a claim if they have overlapping words or similar words. A strong interaction model, known as MatchPyramid\cite{pang2016text,wan2016match,pang2016study,liu2016deep}, is adopted in our baseline model. This technique leverages a similarity matrix plotted from the similarity between a pair of sequences and a CNN with pooling strategies to extract hierarchical interaction patterns. CNNs strength of modeling spatial (position-aware) correlation is utilised for vary length among data. This deep neural network enables us to find the matching patterns between a piece of short text in claim and a long document, which is critical to problems in our task. Multiple layers of 2D convolutions and pooling are used followed by a feed-forward network. \cite{pang2016study} experimented with four similarity functions (indicator function, dot product, cosine and gaussian kernel) and found that using embedding, gaussian kernel similarity function is better than others. A proper kernel size will get more information and generate a better result. Pooling size are used to reduce the dimension of the feature maps, and to pick out the most important information for the latter layers. Especially in ac-hoc retrieval task, documents often contain hundreds of words, but most of them might be background words (exactly same problem in our task). So the pooling layers might be even more important to distill the useful information from the noisy background. 

Inspired by \cite{xie2019visual,vo2020facts}, $q_{text}$ and $d_{text}$ are pre-processed and embedded with pre-trained word embedding model. The embeddings are used to initialise the network. Self-attention layer is applied to embeddings of two inputs since the premise document ($D$) in Factify can be very long and complex. Intuitively, self-attention can be helpful to capture structural information and focus on important keywords particularly in long distance dependencies. Specifically, the scaled dot product (SDP) attention \cite{vaswani2017attention} is used to capture this hidden information: $Attn_{sdp} = softmax(\frac{qK^T}{\sqrt{d_k}})V$, 
$Q_{text\_attn} = Attn_{sdp}(Q_{text})$, $D_{text\_attn} = Attn_{sdp}(D_{text})$, where $q$ here represents $Q$ or $D$, $d_{k}$ denotes the embedding dimension, $Q \in \mathbb{R}^{M \times d_{k}}$ is the claim text ($Q_{text}$) feature matrix and $D \in \mathbb{R}^{N \times d_{k}}$ is the document text ($D_{text}$) feature matrix. $M$ and $N$ is sequence length (of embedding) in matrix $Q$ and $D$. $Attn_{sdp} \in \mathbb{R}^{N \times M}$ computes resulting attention mask for $Q$ and $D$ respectively. 

Subsequently, the self-attended $Q_{text}$ and $D_{text}$ feature matrix is then fed into a GRU layer in order to obtain contextual representation. Finally, dot product function is applied to build similarity matrix between two GRU output sequences for MatchPyramid model in attempt to measure the semantic relevance more accurately between claim and document with higher level of word semantics. The output of MatchPyramid is flattened into a 1D vector and fed into a fully connected multi-layer perceptron (MLP), followed by the Softmax model to perform 3-way classification.


\subsection{5-way Ensemble Model}\label{sec:method_5way_ensemble}
One way to utilize multiple models is to combine uni-modal model predictions in an ensemble classifier to predict final labels. As elaborated in \ref{sec:threeway_text_ent}, 3-way textual entailment model is helpful in distinguishing three-way entailment relationship based on linguistic and semantic clues between $q_{text}$ and $d_{text}$. To address multimodal entailment (as defined in this task \ref{sec:problem_statement}),  a simple relatedness measurement for visual content between $q_{image}$ and $d_{image}$ is adopted based on image pairwise similarity as a proxy of their visual entailment computed with a pre-trained CNN model (ResNet) in our approach. The same as text entailment, this is based on hypothesis and salient correlation patterns observed in this dataset that an article is relevant to a claim if the article contains images similar to the claim’s images.



\begin{figure}[h]
\centering
\includegraphics[width=0.78\linewidth]{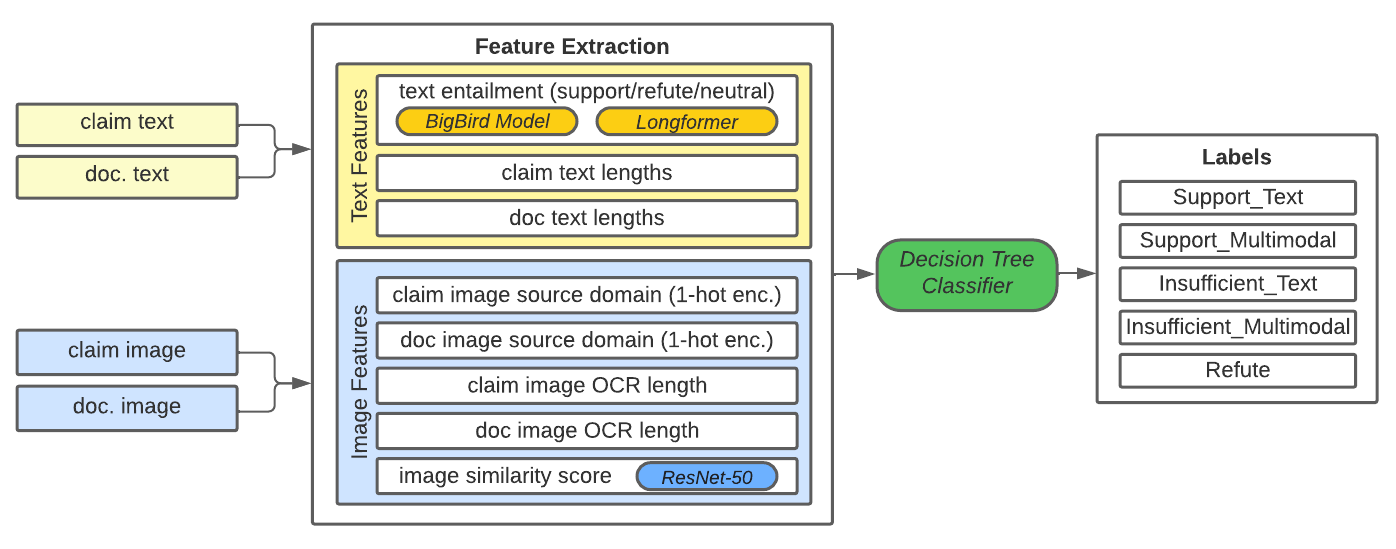}
\caption{Ensemble Model Architecture}
\label{fig:ensemble}
\end{figure}

Hence, an ensemble approach is proposed to combine textual entailment, visual relatedness measurement and additional data specific features. More specifically, the proposed ensemble model utilizes a basic decision tree classifier with the following feature encoding to provide end-to-end five categories classification as depicted in Fig.~\ref{fig:ensemble}.
\begin{itemize}
\item \textit{Length of Text and OCR}: four features of text length between $(D,Q)$ are employed representing the length of $q_{text}$, $OCR(q_{image})$, $d_{text}$, and $OCR(d_{image})$ respectively. OCR text are measured and used as independent features here (ref. \ref{sec:data_doc_len_dist}).
\item \textit{Text Entailment}: two features consisting of a numeric representations ($0$, $1$, $2$) of text entailment prediction (i.e., "insufficient", "support", "refute") along with the corresponding probabilities (Sec.~\ref{sec:threeway_text_ent})
\item \textit{Image Similarity}: the pairwise cosine similarity score between $q_{image}$ and $d_{image}$ is computed based on the features obtained by pre-trained ResNet-50 model,
\item \textit{Image Domain}: two features encoded with one-hot-encoding scheme on the source domain name for $q_{image}$ and $d_{image}$ (ref. \ref{sec:data_image_domain_dist}).
\end{itemize}

\subsection{5-way Multimodal Entailment}
We consider \textit{how to obtain attended multimodal information that can effectively capture \textbf{consistency and integrity} of multimedia content between $D$ and $Q$}. Inspired by current advance in attention techniques \cite{vaswani2017attention,li2019improve,vo2020facts,xie2019visual}, we apply multiple attention mechanisms to learn the multimodal interaction between pairs of visual and textual content. Instead of \textit{local alignment} approaches that model visual objects and textual words, we focus on \textit{global alignment} based methods in this study that aim to map whole image and sentences into joint semantic space. A popular framework to model the multimodal relationship is using a multi-branch attention network, typically one branch projects the image and another models the text. The similarity is measured by dot product on the normalised feature vectors. The extended \textit{MatchPyramid} (elaborated in section \ref{sec:threeway_text_ent}) is applied to model high-level relevance between text pairs. 


In general, our end-to-end multimodal entailment architecture consists of \textit{embedding layer}, \textit{text matching layer}, \textit{multimodal matching layer}, and \textit{classification layer}. Formally, as \textit{input representations}, Images pairs ($q_{image}$ and $d_{image}$) are represented as the top layer (before softmax) of a pre-trained convolutional network and text pairs ($q_{text}$ and $d_{text}$) are mapped into a vector $t \in \mathbb{R}^{j}$ by a fixed word embedding layer initialised by Glove embeddings. $j$ denote as word embedding dimension (e.g., $j$ set to 50 for Glove-50). There is no restriction on the choice of the image encoder but the pre-trained ResNet-50 model is used in our experiments because of its simplicity. 

In \textit{embedding layer},  Let $l$ be the dimension of an image visual vector (i.e., $l=2049$ for ResNet-50) and Let $m$ and $n$ be the number of words in $q_{text}$ and $d_{text}$ respectively. Let $q_{i} \in \mathbb{R}^{l}$ and $q_{t} \in  \mathbb{R}^{j \times m}$ be claim image embedding vector and word embedding matrix, respectively. Likewise, let $d_{i} \in \mathbb{R}^{l}$ and $d_{t} \in \mathbb{R}^{j \times n}$ be document image embedding vector and word embedding matrix, respectively. Each 2048-dim feature vector ($q_{i}$ and $d_{i}$) is fed into a (non-trainable) linear layer to reduce the visual features from 2048 to 512 dimensional vector space in this work. For the embeddings of the text pair, self-attention (SDP) is applied (as specified in section \ref{sec:threeway_text_ent} for both $q_t$ and $d_t$ before feeding into a separate GRU layer to obtain both of their context sequence representation ($q_{t\_cxt} \in  \mathbb{R}^{j \times o}$ and $d_{t\_cxt} \in  \mathbb{R}^{j \times o}$) and corresponding global representation (i.e., final state), denoted by $q_{t\_g} \in \mathbb{R}^{o}$ and $d_{t\_g} \in \mathbb{R}^{o}$. $o$ denotes the GRU output dimension.

In subsequent \textit{text matching} layer, the same pipeline (as specified above for extended MatchPyramid) is applied in attempt to model the high level relevance between article content ($d_{text}$) to claim text ($q_{text}$) based on contextual word embedding interactions. Interaction feature matrix is calculated by the matrix dot product between $q_{t\_cxt}$ and $d_{t\_cxt}$, which is then applied by deep hierarchical convolution layers in the MatchPyramid model to extract aggregated similarity feature vector $Z_{q\_d\_text} \in \mathbb{R}^{f}$ and $f$ is the output dimension of flattened feature maps. The high-level matching patterns are then fed into a multi-layer perceptron (MLP) with dropout to produce the final matching score with learnable weights.

Multimodal latent interaction features are derived in \textit{multimodal matching layer} which mainly consists of \textit{visual matching layer} and \textit{cross-modal attention layer}. Fundamentally, multimodal matching layer aims to find potential relevance of document visual vector ($d_i$) to claim vectors of either visual or text context or both ($q_i$ and $d_{t\_g}$), and are hence critical for predicting multimodal entailment relations for target claim. The same as the process for text embeddings, our \textit{visual matching layer} utilises self-attention (SDP) for image pairs embeddings ($q_{i}$ and $d_{i}$) in attempt to capture important features in each image. Then, a visual similarity feature ($V_{q,d} ^{i} = \norm[2]{\vec{q_{i}}} \cdot \norm[2]{\vec{d_{i}}}$) is computed by applying dot product to the two saliency guided image embeddings with $L2$ norm. This practice follows \cite{malisiewicz2011ensemble,gharbi2012gaussian,xie2019visual}, which is a simple yet very efficient SoTA technique in handling variations in image illumination, view points, texture and season. It also links to feature whitening, linear discrimination analysis and image saliency. In addition, a separate image euclidean similarity feature ($E_{q,d} ^{i} = \dfrac{1}{(1 + \norm[2]{q_{i} - d_{i}})}$) is computed in an attempt to measure potential adversarial images, same people in different scene, cropped image, etc \cite{wang2005euclidean,pedraza2021really}. 

In \textit{cross-modal attention layer}, multiple attention mechanisms are applied, aiming to learn to find and attend to the maximum relevant elements on both feature importance and relational mapping among each self-attended visual and content vectors between claim and document. Specifically, text-image interaction feature between document image and claim text is computed using same SDP attention: $q_{cross\_attn} ^{i,t} = Attn_{sdp}(d_{i}, q_{t\_g})$. Likewise, document image and document text interaction is computed via $d_{cross\_attn} ^{i,t} = Attn_{sdp}(d_{i}, d_{t\_g})$.  Note that due to the mismatched feature space between vision and content, GRU is employed initially to align image features and the text features in order to perform cross-modal learning. Finally, resulting multimodal features $V_{multi} = (V_{q,d} ^{i} \oplus E_{q,d} ^{i} \oplus q_{cross\_attn}^{i,t} \oplus d_{cross\_attn} ^{i,t})$ are obtained from the merge (with concatenation) of the two intermediate layers output before feeding into a MLP with dropout to generate fused higher level features. 

Finally, in the \textit{classification layer}, the outputs of two MLP layers are merged along with corresponding hypothesis multimodal representation ($q_{merge}$) into a combined representation with batch normalisation. $q_{merge}$ is the output of a separate MLP layer for merged (concatenation) $q_{t\_g}$ and normalised $q\_{i}$ ($q_{merge} \in \mathbb{R}^{o+l}$). Final representation are applied with dropout regularization before feeding to a softmax layer to output the probabilities of five categories.




\section{Factify Dataset}\label{sec:data}

\subsection{Data statistics}

Five entailment categories are balanced in both train and validation data set. As mentioned above, both claim (\textit{hypothesis}) and document(\textit{premise}) contains an image and a text of vary length. Optical character recognition (OCR) text extracted from image are not counted separately here and combined with corresponding claim and document text respectively. Thus, large claim word size presented in the table is due to extracted OCR text. The data details for each set are shown in Table \ref{tab:data}. More details about the dataset and task details can be referred in \cite{mishraFactify2021}.


\begin{table}
 \caption{Data statistics}\label{tab:data}
 \small\centering
 \begin{tabular}{lS}
    \toprule
     \textbf{Data set sizes:} & \\
     Train pairs: &   35,000 \\
     Validation pairs:  & 7,500 \\
     Test pairs:      &   7,500 \\
     \midrule
     \textbf{Sentence length}: &\\
     Claim (Hypotheis) inc. OCR & \\
     \multicolumn{1}{r}{min token count:} & \multicolumn{1}{r}{1}\\
     \multicolumn{1}{r}{max token count:} & \multicolumn{1}{r}{19,105} \\
     \multicolumn{1}{r}{mean token count:} & \multicolumn{1}{r}{51.5}\\
     Document(Premise) inc. OCR &\\
     \multicolumn{1}{r}{min token count:} & \multicolumn{1}{r}{1}\\
     \multicolumn{1}{r}{max token count:} & \multicolumn{1}{r}{44,542}\\
     \multicolumn{1}{r}{mean token count:} & \multicolumn{1}{r}{1010.5}\\
     \bottomrule
 \end{tabular}
\end{table}

\subsection{Word Overlaps distribution}
Word overlap is an important indication for modeling textual entailment as well as of potential data bias. Naturally, when pairing claims to evidence sentences, the word overlap ratio will be higher on average for claims with their supporting evidence. However, models relying on word-overlap will perform poorly when dealing with complexity in the real world examples (typically antonymous examples and adversarial attacks). In VITAMINC\cite{schuster2021get} and FEVER\cite{thorne2020evaluating} dataset, this bias is deliberately minimized in order to create challenging examples that require sentence-pair inference and cannot be solved by simple word matching techniques. Here, the word overlaps distribution per class in train and val set are presented in Table \ref{tab:train_wordoverlap}. The data distribution indicates that evidential premise data in Factify have clearly high word overlap ratio than other two categories of insufficient evidence.


\begin{table}[h!]
  \begin{center}
    \caption{${(Q,D)}$ pair text word overlap dist. in train/val set}\label{tab:train_wordoverlap}
    \begin{tabular}{l|S|r|r|r}
      \textbf{Category} & \textbf{min.} & \textbf{max.} & \textbf{mean} & \textbf{mdn.}\\
      \hline
      Support\_Multi. & 0.0 & 1.0 & 0.299 & 0.273 \\
      Support\_Text & 0.0 & 1.0 & 0.316 & 0.294\\
      Insufficient\_Multi. & 0.0 & 0.92 & 0.221 & 0.192\\
      Insufficient\_Text & 0.0 & 1.0 & 0.238 & 0.176 \\ 
      Refute & 0.0 & 1.0 & 0.406 & 0.346 \\ 
    \end{tabular}
  \end{center}
\end{table}


\subsection{Image Similarity distribution}
Image similarity is the most basic indicator of multimodal entailment. To empirically validate the intuition of potential similarity correlation between $q_{image}$ and $d_{image}$ (as mentioned in \ref{sec:method_5way_ensemble}) in the dataset, image relatedness analysis is conducted in this section. Similar to the potential bias from data overlapping, we compute image pairwise similarity distribution with embedding space computed from pre-trained ResNet50 model over train/val set as presented in Table \ref{tab:train_imagesim_dist} and \ref{tab:val_imagesim_dist}. As seen from the distribution over five categories, two text related entailment categories have clearly lower pairwise image similarity that multimodal evidential entailment categories.

\begin{table}[h!]
  \begin{center}
    \caption{${(Q,D)}$ image pairwise similarity dist. in train set}\label{tab:train_imagesim_dist}
    \begin{tabular}{l|S|r|r|r}
      \textbf{Category} & \textbf{min.} & \textbf{max.} & \textbf{mean} & \textbf{mdn.}\\
      \hline
      Support\_Multi. & 0.533 & 1.0 & 0.864 & 0.865 \\
      Support\_Text & 0.327 & 1.0 & 0.704 & 0.725 \\
      Insufficient\_Multi. & 0.428 & 0.999 & 0.835 & 0.833\\
      Insufficient\_Text & 0.408 & 0.971 & 0.703 & 0.722 \\ 
      Refute & 0.41 & 1.0 & 0.82 & 0.835 \\ 
    \end{tabular}
  \end{center}
\end{table}

\begin{table}[h!]
  \begin{center}
    \caption{${(Q,D)}$ image pairwise similarity dist. in val set}
    \label{tab:val_imagesim_dist}
    \begin{tabular}{l|S|r|r|r}
      \textbf{Category} & \textbf{min.} & \textbf{max.} & \textbf{mean} & \textbf{mdn.}\\
      \hline
      Support\_Multi. & 0.533 & 1.0 & 0.855 & 0.856 \\
      Support\_Text & 0.393 & 1.0 & 0.72 & 0.74 \\
      Insufficient\_Multi. & 0.578 & 0.996 & 0.846 & 0.844\\
      Insufficient\_Text & 0.383 & 0.936 & 0.71 & 0.73 \\ 
      Refute & 0.426 & 1.0 & 0.828 & 0.842 \\ 
    \end{tabular}
  \end{center}
\end{table}

\subsection{Text Length Distribution}\label{sec:data_doc_len_dist}
\begin{figure}[h]
\centering
\includegraphics[width=.40\linewidth]{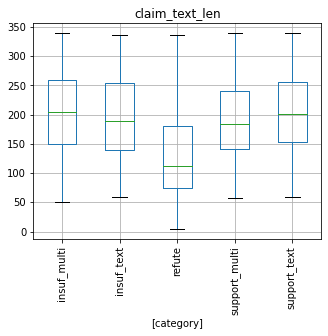}\hspace{.2cm}
\includegraphics[width=.41\linewidth]{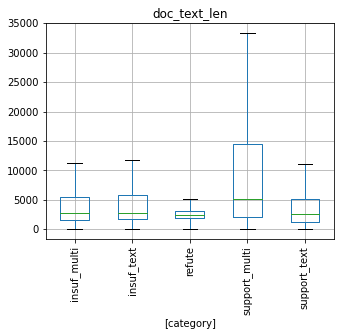}
\caption{Text length distribution over the five entailment categories}
\label{fig:text_len_dist}
\end{figure}

\begin{figure}[h]
\centering
\includegraphics[width=.39\linewidth]{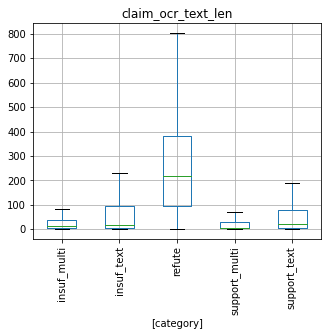}\hspace{.2cm}
\includegraphics[width=.4\linewidth]{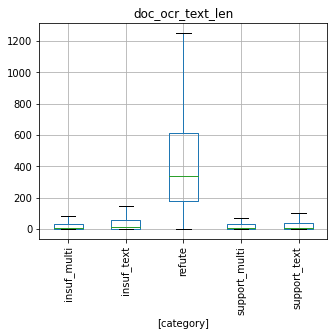}
\caption{OCR text length distribution over the five entailment categories}
\label{fig:ocr_len_dist}
\end{figure}

Text and OCR (text) length distributions between $(Q,D)$ in the train set are presented in Fig~\ref{fig:text_len_dist} and Fig~\ref{fig:ocr_len_dist}. Clearly separable distribution patterns can be seen across five categories in claim and document text and their corresponding OCR text. As shown in Fig~\ref{fig:text_len_dist}, document text length varies most for 'Support\_Multimodal' among other four entailment categories. The document length in two insufficient categories share similar range and 'Refute' category has the least document length. The claim length distribution shows a clear bias of 'Refute' examples towards shorter claims. While the remaining classes present similar ranges, 'Insuffient\_Text' and 'Support\_Multimodal' tend to include slightly shorter claims. In comparison, OCR text for both claim and document in "Refute" category samples shows surprising longer length than other four categories. Motivated by the observation, we adopted text lengths as features in our ensemble model as illustrated in \ref{sec:method_5way_ensemble}.

\subsection{Image Domain Bias}\label{sec:data_image_domain_dist} 
\begin{figure}[h]
\includegraphics[scale=0.30]{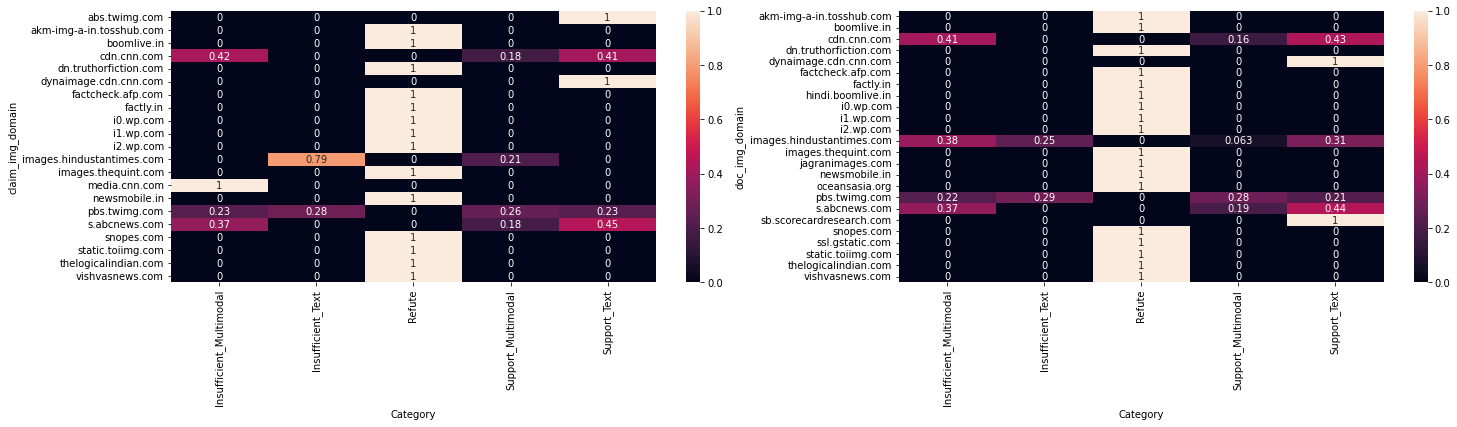}
\caption{Label Distribution by Image Domain in claim and document}
\label{fig:doc_img_domain_dist}
\end{figure}

Source bias is one of the known and common problems in machine learning dataset \cite{torralba2011unbiased} which happens when most data samples are collected from the same source. This problem has several links which mainly include selection bias, capture bias (bias happens due to particular data collection methods), label bias and negative set bias. We are interested in probing potential source bias in Factify datset. The potential bias in images domains over five categories brought to our attention in data analysis. This is important since multimedia metadata have been proved to be valuable information and signal for fact verification \cite{chen2019deep,zlatkova2019fact,prabhakar2021check} in real-world application such as domain/source credibility, detection of image manipulation and tampering. Image (link) domains are extracted from all document samples of train set and distributions are computed across five categories. As shown in Fig.~\ref{fig:doc_img_domain_dist}, experiments show a surprisingly strong correlation between image domains and each entailment categories in both claim and document. Motivated by the correlation analysis, image domains are employed as features in our ensemble model as illustrated in section \ref{sec:method_5way_ensemble}.


\section{Experiments}\label{sec:experiment}

\subsection{Experimental Setting}
To evaluate the performance of our baseline solutions, we use weighted average F1 for benchmarking on validation set. All the experiments are implemented on a single NVIDIA A100 GPU with up to 20 GiB RAM.


\subsection{Hypothesis Only Test}\label{sec:hypo_only_test}
We conducted hypothesis only reliance test by using hypothesis only information to train a model as baseline. This is a commonly adopted approach  \cite{gururangan2018annotation,vu2018grounded} in SNLI/RTE to verify the presence of data bias. The assumption is that without any premise information, this baseline is supposed to make a random guess out of the five classes. We train two models with or without images and test the resulting accuracy for each model on val set.

Two models (Hypo${_{text}}$ and Hypo${_{text+img}}$) are implemented with similar architecture that consists of a text processing component and/or ResNet embedding layer followed by two fully-connected (FC) layers. The text processing component is used to extract the text feature from the given hypothesis. It firstly generates a sequence of word embeddings for the given claim text. The embedding sequence is then fed into a GRU \cite{chung2014empirical} to output the text context features of dimension 300. The image processing component involves BGR to RGB conversion, resizing images (to [300, 300, 3]), the feature extraction with ResNet50 with a linear layer projecting pre-trained embeddings to 512 dimensional vector. The input and output dimensions of two FC layers for text only models are [300, 300] and [300, 3] respectively. For text+image model, the hidden layers dimensions are [300, 300] and [300, 3] respectively.

Our experiments shows that the resulting accuracy and weighted F1 achieves same value in both 0.60 and both 0.64 on val set for text and text+image models respectively, implying the existence of bias in Factify dataset. The details are presented in results section.  Our proposed solutions outperform the two hypothesis only baselines.

\subsection{3-way text entailment}
\textbf{Transformer fine-tune settings: } For the best entailment model, the pre-trained BigBird model from Huggingface and the implementation for pair-wise classification fine-tune was used. For our experiment the model was fine-tuned for 2 epochs, using the AdamW optimizer with learning rate 2e-5 and epsilon 1e-8 with batch size 4. Maximum sentence length was set to the mean length of the input texts namely, 1396 tokens. To train the 3-way entailment model, the 5-way data categories were converted to 3-way categories including "Support" ("Support\_Multimodal"+"Support\_Text"), "Refute", Insufficient("Insufficient\_Multimodal"+"Insufficient\_Text"). OCR text from both $Q$ and $D$ were excluded.

\textbf{MatchPyramid baseline model settings:} We use Glove embeddings with 50 dimension (GloVe 6B 50d) for text input and GRU output dimension is set to 50. The number of CNN layers is set to 2, each with kernel size $3 \times 3 $, pooling size $5 \times 10$ and in valid mode (i.e., no padding). ReLu activation is placed between all convolutional layers. Convolution channels are set to 16 and 32 respectively. 2-layer MLP are set with hidden dimensions of 128 and 64 respectively. The maximum text length for $Q_{text}$ and $D_{text}$ are set to 100 and 1000 respectively.  Limited experiments are conducted including content context size of claim and document, global average pooling and convolution padding schemes.

\subsection{5-way Ensemble Model}

The ensemble model is implemented using \texttt{sklearn}'s DecisionTreeClassifier class. Best pre-trained transformer ("BigBird") based text entailment classifier is adopted and pairwise image similarity is computed based on ResNet-50. For image pre-processing and feature extraction, same practice introduced in \ref{sec:hypo_only_test} is applied. One-hot encoding function provided in Scikit-learn is utilised to converts two categorical features (i.e., URL domains of $q_{image}$ and $d_{image}$) as a one-hot numeric array learnt from train set. Text are pre-processed separately for BigBird model and ensemble model. No pre-processing is applied for four text length features. For the ensemble model training, we use ‘best' split based on ‘gini' impurity matrix as training criteria and limit the number of layers to $8$ to avoid overfitting.


\subsection{5-way Multimodal classification}
5-way end-to-end Multimodal Entailment model \textit{Multimodal${_{ent}}$} is implemented with Keras and tensorflow(v2.4) with Adam optimiser with an adaptive learning scheduler. The initial learning rate and weight decay are both set to 0.0001. Batch size is 32 and maximum number of training epochs is set to 80. Optimal parameters and settings from MatchPyramid baseline model experiments are applied. Checkpoint callback is used to save best model that achieves best validation accuracy. ReLu activation is applied to all convolution layers and fully-connected layers. The uniform He initialization ("he\_uniform") is used for all ReLU layers. Same settings (layer size, hidden dim, activation, etc) are applied for three separate MLP layers. Few parameters and architectures are experimented including vary lengths of claim and document content, MLP layer size (1-2), hidden layer dimensions of MLPs(64,256,512,768,1024), merge strategies (concatenation and multiplication) of three MLPs output for classification layer. Ablation study is conducted by removing individual sub-components including hypothesis MLP ($q_{merge}$), document crossmodal interaction($d_{cross\_attn} ^{i,t}$), and document-claim crossmodal interaction($q_{cross\_attn}^{i,t}$). The ablation experiments show the effectiveness of full model architecture. Best model (as reported in result \ref{res:5wayclass}) is obtained at training epoch 9 and stopped at epoch 14 with optimal settings of the 3 MLPs architecture, 1-layer MLP with dimension 256, MLPs outputs merged with concatenation, and text input lengths of $q_{text}$ and $d_{text}$ with 100 and 1000 respectively.


\section{Results and Discussion}\label{sec:results}
\subsection{3-way text entailment}

The results of the 3-way text entailment models are presented in Table \ref{tab:3wayresults_text}. To validate our model choice, we evaluate few SoTA pre-trained transformer models, including BERT, RoBerta, BigBird and LongFormer. The best performing models are BigBird and LongFormer with the overall winner being BigBird because of the slightly better results and smaller input context size required (1396 vs 1484 respectively). 
\begin{table*}
 \caption{Three-way Classification Results on val set}\label{tab:3wayresults_text}
 \small\centering
 \begin{tabular}{lSSSSSSSSS}
    \toprule
    \multirow{2}{*}{Categories} &
     \multicolumn{3}{|c}{$MatchPyramid_{glove50d}$} &
     \multicolumn{3}{|c}{BigBird} &
     \multicolumn{3}{|c}{LongFormer}\\
     & {P} & {R} & {F1} & {P} & {R} & {F1} & {P} & {R} & {F1}\\
     \midrule
    Support & 0.77 & 0.74 & 0.76 & 0.83 & 0.86 & \textbf{0.85} & 0.83 & 0.86 & 0.84 \\
    Refute & 0.99 & 0.99 & 0.99 & 1.00 & 1.00 & \textbf{1.00} & 1.00 & 1.00 & \textbf{1.00} \\
    Insufficient & 0.75 & 0.77 & 0.76 & 0.85 & 0.83 & \textbf{0.84} & 0.85 & 0.82 & \textbf{0.84} \\
    \midrule
    Weighted Avg. & 0.81 & 0.81 & 0.81 & 0.88 & 0.87 & \textbf{0.87} & 0.87 & 0.87 & \textbf{0.87} \\
    \bottomrule
 \end{tabular}
\end{table*}

For the architecture of extended MatchPyramid baseline, we have experimented with different parameters such as longer context length in $Q_{text}$ (including 1500, 2000, 3000), Glove model with 300 dimension, larger GRU output dimension of 300, various pooling size ([3, 10]), etc, none of these attempts provide major improvement. Overall, our baseline implementation with self-attention and Glove-50d based contextual representation learning achieve optimal performance, which is competitive  to large transformer model based approaches as presented in Table \ref{tab:3wayresults_text}.

\subsection{5-way classification}\label{res:5wayclass}

The results of 5-way $Ensemble$ and $Multimodal_{ent}$ model on val set are presented in Table \ref{tab:5wayresults}. Four of our baseline methods outperform all baseline models proposed by task organiser as reported in  \cite{mishraFactify2021}. The result of best baseline model (Multimodal${_{factify}}$) in Factify data paper is presented in the table \footnote{the corresponding class-wise performance are not provided by organisers}. Unsurprisingly, our ensemble model achieved best results on val set with 0.77 F1 which is 8\% higher than the results of $Multimodal_{ent}$ model. The experiment results demonstrate a large performance gain with the large pre-trained text entailment model which works effectively on long paragraphs and contribute the most towards predicting final 5-way categories. This is particularly obvious for "Refute" label, the samples in which are mostly relying on text based inference.  It is not surprising that the useful features incorporated from the heuristics and bias learned from the dataset have proved to be effective for this multimodal prediction task. 

\begin{table}
 \caption{5-way Classification Results on val set}\label{tab:5wayresults}
 \small
 \resizebox{\columnwidth}{!}{
 \begin{tabular}{lSSSSSSSSSSSSS}
    \toprule
    \multirow{1}{*}{Categories} &
     \multicolumn{1}{|c}{Multimodal${_{factify}}$} &
     \multicolumn{3}{|c}{Hypo${_{text}}$} &
     \multicolumn{3}{|c}{Hypo${_{text+img}}$} &
     \multicolumn{3}{|c}{Multimodal${_{ent}}$} &
     \multicolumn{3}{|c}{Ensemble} \\
     & {F1} & {P} & {R} & {F1} & {P} & {R} & {F1} & {P} & {R} & {F1} & {P} & {R} & {F1}\\
     \midrule
    Support\_Multimodal & n/a & 0.60 & 0.60 & 0.60 & 0.59 & 0.63 & 0.61 & \textbf{0.84} & 0.57 & 0.68 & 0.74 & 0.78 & \textbf{0.76}\\
    Support\_Text & n/a & 0.48 & 0.43 & 0.45 & 0.55 & 0.51 & 0.53 & 0.51 & 0.66 & 0.58 & 0.71 & 0.71 & \textbf{0.71}\\
    Insufficient\_Multimodal & n/a & 0.45 & 0.52 & 0.61 & 0.56 & 0.61 & 0.56 & 0.62 & 0.52 & 0.57 & 0.68 & 0.65 & \textbf{0.66} \\
    Insufficient\_Text & n/a & 0.61 & 0.50 & 0.55 & 0.62 & 0.51 & 0.56 & 0.57 & 0.69 & 0.62 & 0.74 & 0.73 & \textbf{0.73}\\
    Refute & n/a & 0.87 & 0.90 & 0.88 & 0.93 & 0.93 & 0.93  & 0.99 & 0.97 & 0.98 & 1.0 & 1.0 & \textbf{1.0} \\
    \midrule
    Weighted Avg.& 0.54 & 0.60 & 0.60 & 0.60 & 0.64 & 0.64 & 0.64 & 0.71 & 0.68 & 0.69 & 0.77 & 0.77 & \textbf{0.77}\\
    \bottomrule
 \end{tabular}
 }
\end{table}

It was found that differentiating between \textit{"Insufficient\_Multimodal"} and \textit{"Support\_Text"} or between \textit{"Insufficient\_Text"} and \textit{"Support\_Multimodal"} was the most challenging task without relying data specific features. In other words, when a sample contains supporting document text for the claim but the image is irrelevant, our model has low confidence in predicting the label as \textit{"Support\_Text"} or \textit{"Insufficient\_Multimodal"}. Likewise, when document image is relevant to claim image about same information context but document text is irrelevant, our model has low confidence in predicting the correct label.  The decision is highly dependent on the annotation bias. From all the labels, the "Refute" label is the most distinguishable category and highly dependent on the text. The performance is highly consistent among all our models and participant systems in this competition (as seen in leaderboard \ref{tab:factify_leaderboard}). This is possibly mainly attributed to the articles samples selected from very few fact checking sources that have highly differentiable linguistic clues (typically high frequent negative words used and same verdict sentences frequently appeared in this category such as "The claim is false"). 



\subsection{Competition Result}
Final test set results and competition leaderboard are presented in Table \ref{tab:5wayresults_test} and \ref{tab:factify_leaderboard} respectively. Our best model ("Ensemble") outperform all competition systems and best baseline models. Test result of $Ensemble$ model achieved 0.77 avg. F1 which is the same as the result on val set and 10\% higher than the result of $Multimodal_{ent}$.

\begin{table}
 \caption{5-way Classification Results on test set}\label{tab:5wayresults_test}
 \small
 \begin{tabular}{lSSSSSS}
    \toprule
    \multirow{1}{*}{Categories} &
     \multicolumn{3}{|c}{Multimodal${_{ent}}$} &
     \multicolumn{3}{|c}{Ensemble} \\
      & {P} & {R} & {F1} & {P} & {R} & {F1}\\
     \midrule
    Support\_Multimodal & 0.81 & 0.60 & 0.69 & 0.76 & 0.78 & \textbf{0.77}\\
    Support\_Text  & 0.47 & 0.59 & 0.52 & 0.65 & 0.69 & \textbf{0.67}\\
    Insufficient\_Multimodal & 0.61 & 0.53 & 0.57 & 0.73 & 0.64 & \textbf{0.68} \\
    Insufficient\_Text  & 0.56 & 0.66 & 0.60 & 0.71 & 0.73 & \textbf{0.72} \\
    Refute  & 0.99 & 0.96 & 0.98 & 1.0 & 1.0 & \textbf{1.0} \\
    \midrule
    Weighted Avg. & 0.69 & 0.67 & 0.67 & 0.77 & 0.77 & \textbf{0.77}\\
    \bottomrule
 \end{tabular}
\end{table}

\begin{table}
 \caption{Factify Official Leaderboard}\label{tab:factify_leaderboard}
 \small
 \resizebox{\columnwidth}{!}{
 \begin{tabular}{lSSSSSSS}
    \toprule
     \multicolumn{1}{|c}{Rank} &
     \multicolumn{1}{|c}{Team} &
     \multicolumn{1}{|c}{Support\_Text} &
     \multicolumn{1}{|c}{Support\_Multi.} &
     \multicolumn{1}{|c}{Insufficient\_Text} &
     \multicolumn{1}{|c}{Insufficient\_Multi.} &
     \multicolumn{1}{|c}{Refute} &
     \multicolumn{1}{|c}{Final}
     \\
     \midrule
   1 & Logically & 81.843\% & 87.429\% & 84.437\% & 78.345\% & 99.899\% & \textbf{76.819\%} \\
   2 & {Yet}       & 75.518\% & 89.38 \%  & 82.121\% & 80.81\%  & 99.866\% & 75.591\% \\
   3 & {Truthformers} & 77.65\% & 85.057\% & 79.421\% & 84.482\% & 98.819\% & 74.862\% \\
   4 & {UofA-Truth}  & 78.493\% & \textbf{89.786\%} & 82.995\% & 75.981\% & 98.339\% & 74.807\% \\
   5 & {Yao} & 68.881\% & 81.61\% & 84.836\% & \textbf{88.309\%} & \textbf{100.0\%} & 74.585\% \\
   6 & {Greeny} & 74.947\% & 86.018\% & 80.382\% & 82.858\% & 99.125\% & 74.28\% \\
   7 & {GPTs} & 71.575\% & 79.032\% & 75.363\% & 79.275\% & \textbf{100.0\%} & 69.461\% \\
   8 & {Tyche} & 75.0\% & 75.259\% & \textbf{85.496\%} & 68.823\% & 99.159\% & 69.203\% \\
   9 & {MUM\_NLP} & 64.803\% & 80.857\% & 69.848\% & 66.548\% & 93.465\% & 61.165\% \\
   \midrule
    - & {BASELINE} & \textbf{82.675\%} & 75.466\% & 74.424\% & 69.678\% & 42.354\% & 53.098\% \\
    \bottomrule
 \end{tabular}}
\end{table}


\section{Conclusion}\label{sec:conclude}
We described our participation in the Multimodal Fact Verification Factify Challenge with the implementation of two proposed baseline solutions including an ensemble model and an end-to-end multimodal entailment model. Ensemble model based system outperform the end-to-end model on val set and test set. The best performing model in this competition combines results of 3-way text entailment classifier, visual similarity with a pre-trained CNN model and heuristics learnt from the dataset. Multimodal fusion technique is explored in this paper to model interaction between different modalities (i.e., text and image) in claim and document pairs and combines information from them to learn multimodal entailment relationship end-to-end. We found that multimodal entailment based system suffer from overfitting. Apart from limited train size and identified data bias, our experiments suggest that fine-grained image and text interaction model need to be explored further. 

We found that the ambiguous labels in Factify dataset undermines the performance of our deep learning architecture. Creating a dataset for a complex real-word multimodal NLP problems particularly natural language inference as multimodal verification has raised emergent challenges \cite{le2020adversarial,sharma2021evaluating} and indeed a cumbersome task, and we appreciate the work by the Factify organizers, yet, a more elaborate and unbiased dataset along with well defined annotation criterion should make this dataset more suitable for benchmark. More effort is required to tackle the dataset challenge of minimising hypotheses from human annotators and make dataset better reflecting real-world challenges. As an emergent research field, we hope our extensive data analysis and proposed baseline solutions can inspire further work.









\bibliography{factify}




\end{document}